\documentclass[10pt,twocolumn,letterpaper]{article}

\usepackage{iccv}
\usepackage{times}
\usepackage{epsfig}
\usepackage{graphicx}
\usepackage{amsmath}
\usepackage{amssymb}

\usepackage{multirow}
\usepackage[table,xcdraw]{xcolor}

\usepackage[breaklinks=true,bookmarks=false]{hyperref}

\iccvfinalcopy % *** Uncomment this line for the final submission

\def\iccvPaperID{1765} % *** Enter the ICCV Paper ID here

\ificcvfinal\fi

\begin{document}

%%%%%%%%% TITLE
\title{PointMBF: A Multi-scale Bidirectional Fusion Network for Unsupervised RGB-D Point Cloud Registration}

\author{
{Mingzhi Yuan$^{1,2}$, Kexue Fu$^{3\thanks{Equal contribution.}}$, Zhihao Li$^{1,2}$, Yucong 
Meng$^{1,2}$ and Manning Wang$^{1,2\thanks{Corresponding author.}}$} \\
$^{1}$Digital Medical Research Center, School of Basic Medical Sciences, Fudan University, China\\
$^{2}$Shanghai Key Laboratory of Medical Image Computing and Computer Assisted Intervention, China\\
$^{3}$Shandong Computer Science Center (National Supercomputer Center in Jinan)
\\
{\tt\small \{mzyuan20,fukexue,mnwang\}@fudan.edu.cn,  \{lizhihao21,ycmeng21\}@m.fudan.edu.cn }
}

\maketitle
\ificcvfinal\thispagestyle{empty}\fi

%%%%%%%%% ABSTRACT
\begin{abstract}
   Point cloud registration is a task to estimate the rigid transformation between two unaligned scans, which plays an important role in many computer vision applications. 
   Previous learning-based works commonly focus on supervised registration, which have limitations in practice. 
   Recently, with the advance of inexpensive RGB-D sensors, several learning-based works utilize RGB-D data to achieve unsupervised registration. 
   However, most of existing unsupervised methods follow a cascaded design or fuse RGB-D data in a unidirectional manner, which do not fully exploit the complementary information in the RGB-D data. 
   To leverage the complementary information more effectively, we propose a network implementing multi-scale bidirectional fusion between RGB images and point clouds generated from depth images. 
    By bidirectionally fusing visual and geometric features in multi-scales, more distinctive deep features for correspondence estimation can be obtained, making our registration more accurate. 
   Extensive experiments on ScanNet and 3DMatch demonstrate that our method achieves new state-of-the-art performance. 
   Code will be released at \url{https://github.com/phdymz/PointMBF}.  
 \end{abstract}
 
 %%%%%%%%% BODY TEXT
 \section{Introduction}
 Point cloud registration \cite{ref1} aims at aligning partial views of the same scene, which is a critical component of many computer vision tasks. 
 Commonly, point cloud registration starts from feature extraction \cite{ref2,ref3} and correspondence estimation \cite{ref4,ref5}, followed by robust geometric fitting \cite{ref6,ref7,ref8,ref64}. 
%  Among them, feature extraction plays a vital role in point cloud registration, as distinctive features can avoid outlier correspondences and save time on robust geometric fitting. 
Among them, feature extraction plays a vital role in point cloud registration, as distinctive features can reduce the occurrence of outlier correspondences, thereby saving time on robust geometric fitting.

 Many traditional methods rely on hand-crafted features \cite{ref2,ref9}, but they commonly show limited performance. 
 Benefiting from the rapid progress of deep learning, many learning-based features \cite{ref3,ref10,ref11} have been proposed in recent years. 
 Compared to hand-crafted features, they are distinctive enough to achieve robust performance in many challenging conditions such as low overlap. 
 However, most deep learning-based features need supervision on poses or correspondences, which limits their practical applications. 
 For unannotated datasets with different distributions from the training set, they tend to suffer from performance degradation.
 
 With the recent advance of inexpensive RGB-D sensors, it has become easier to simultaneously acquire both depth information and RGB images, which inspires unsupervised point cloud registration using additional color information. 
 UR\&R \cite{ref12} proposed a framework for unsupervised RGB-D point cloud registration. 
 It utilizes a differentiable renderer to generate the projections of the transformed point clouds and calculates geometric and photometric losses between the projections and the registration targets. 
 Based on these losses, UR\&R can train its deep descriptor without annotations and achieve robust registration on RGB-D video. 
 Similar to UR\&R, BYOC \cite{ref13} proposed a teacher-student framework for unsupervised point cloud registration for RGB-D data, which also shows competitive performance. 
 However, all these RGB-D-based methods use RGB images and depth information separately and do not further exploit the complementary information within RGB-D data. 
 Recently, LLT \cite{ref14} first utilized a linear transformer \cite{ref15,ref16} to fuse these complementary information and achieved new state-of-the-art performance. 
 However, LLT focuses on using depth information to guide RGB information and neglects the interaction between the two modalities, which hinders better performance. 
 
 To fully leverage these complementary modalities, we propose a \textbf{m}ulti-scale \textbf{b}idirectional \textbf{f}usion network named PointMBF for unsupervised RGB-D point cloud registration, which fuses visual and geometric information bidirectionally at both low and high levels. 
 In this work, we process depth images in the form of point clouds and utilize two network branches for RGB images and point clouds, respectively. 
 Both branches follow the U-Shape \cite{ref18} structure to extract features for information fusion in multiple scales. 
 Unlike the fusion strategy in LLT \cite{ref14}, we perform cross-modalities fusion in all stages rather than only in the last few layers, making fused features more distinctive. 
 Moreover, different from the unidirectional fusion strategy in LLT, we adopt a bidirectional design for more effective fusion. 
 % Specifically, in each scale, we first sample the KNN points for each query pixel in its corresponding point neighbor and vice versa for each query point. 
 % Then we utilize a PointNet-style \cite{ref20} module to aggregate the features of samples. 
 % Finally, we bridge the information communication between different modalities by concatenating the aggregated features with the query feature and mapping them through a multi-layer perceptron. 
 % By this multi-scale bidirectional fusion in feature extraction, we can obtain more distinctive features for correspondence estimation and all above modules can be unsupervisedly trained by a differentiable render. 
 Specifically, in each scale, we first find the regional corresponding points/pixels for each query pixel/point. 
 Then we sample the KNN points/pixels among them and gather their features to a set. 
 The feature set is fed to a PointNet-style module to achieve permutation-invariant aggregation. 
 Finally, the information communication between different modalities can be achieved by fusing the aggregated features with the query feature using a shallow neural network with residue design.

 To evaluate our method, we conduct experiments on two popular indoor RGB-D datasets, ScanNet \cite{ref21} and 3DMatch \cite{ref22}. 
 Our PointMBF not only achieves new state-of-the-art performance but also shows competitive generalization across different datasets. 
 When tested on an unseen dataset ScanNet, our PointMBF trained on 3DMatch still shows comparable performance to recent advanced methods directly trained on ScanNet. 
 % When tested on an unseen dataset, it still shows comparable performance to recent advanced methods trained by data consistent with the unseen dataset. 
 We also conduct comprehensive ablation studies to further demonstrate the effectiveness of each component of our multi-scale bidirectional design.
 
 To summarize, our contributions are as follows: 
 \begin{itemize}
   \item We propose a multi-scale bidirectional fusion network for RGB-D point cloud registration, which fully leverages the information in the two complementary modalities. Compared to unidirectional fusion or fusion in the final stage, our fusion strategy can achieve the information communication more effectively, so that it can generate more distinctive features for registration. 
   \item We introduce a simple but effective module for bidirectional fusion, which adapts to density-variant point clouds generated by view-variant depth images. 
   \item We provide a comprehensive comparison between different fusion strategies to analyze their effect empirically.
   \item Our method achieves new state-of-the-art results on RGB-D point cloud registration on ScanNet \cite{ref21} using weights trained either on ScanNet or 3DMatch \cite{ref22}.
 \end{itemize}
 
 \begin{figure*}[ht]
   \begin{center}
   \centerline{\includegraphics[width=2.0\columnwidth]{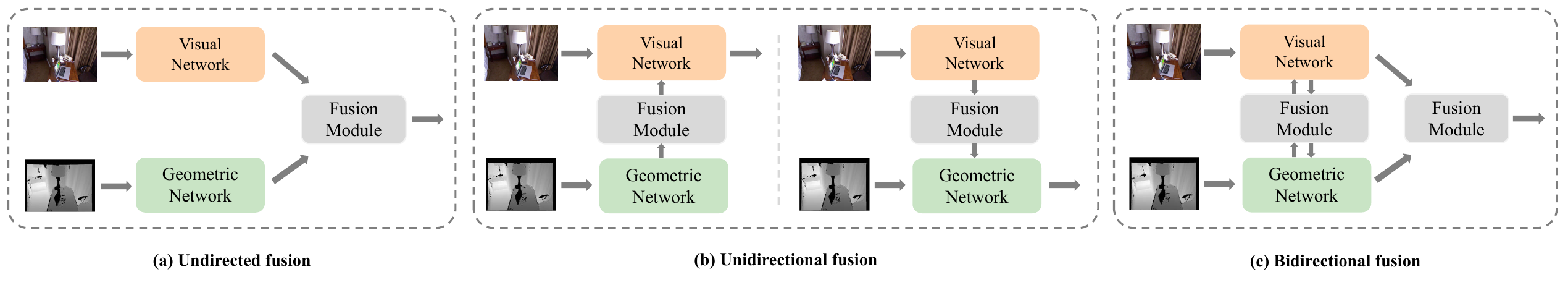}}
   \caption{
     Comparison on fusion strategies.
     }
   \label{fig1}
   \end{center}
   % \vspace{-15pt}
 \end{figure*}

 \section{Related Work}
 \subsection{Point Cloud Registration}
 Point cloud registration aims at aligning partial scan fragments, which is widely used in many tasks such as autonomous driving \cite{ref44}, robotics \cite{ref43}, and SLAM \cite{ref45}. 
 Except for some ICP-based methods \cite{ref42,ref46}, metric-based methods \cite{ref47,ref48,ref59}, and so on, most methods follow the process of feature extraction \cite{ref2,ref3,ref65}, correspondence estimation \cite{ref4,ref5}, and robust geometric fitting \cite{ref6,ref7,ref8}. 
 In the past, many traditional methods were often limited by hand-crafted features \cite{ref2,ref9}. 
 % In some applications, the hand-crafted 3D descriptor is not distinctive enough to generate a certain number of inlier correspondences, so estimations are still unreliable even under time-consuming geometric fitting methods such as RANSAC \cite{ref6}. 
 Recently, many learning-based 3D descriptors \cite{ref3,ref39,ref40,ref41,ref38,ref49} were proposed. 
 They have achieved impressive performance and some methods \cite{ref38,ref49} are even free of RANSAC. 
 However, most of them rely on pose or correspondence supervision, which limits their practical application. 
 For unannotated datasets, they can only infer using weights trained on other datasets, which tends to degrade their performance. 
 Benefiting from inexpensive RGB-D sensors, many RGB-D video datasets \cite{ref21,ref22} were proposed. 
 The extra color information contains richer semantics and many works achieve unsupervised learning based on it. 
 To the best of our knowledge, UR\&R \cite{ref12} is the first learning-based work using RGB-D data for unsupervised registration. 
 It also follows the above mentioned registration process but it utilizes a differentiable renderer-based loss to optimize its learnable descriptor. Inspired by self-supervised learning \cite{ref50}, BYOC \cite{ref13} proposed a teacher-student framework for 3D descriptor unsupervised learning. 
 It teaches a 3D descriptor by a 2D descriptor, making the 3D descriptor achieve comparable performance to supervised methods. 
 However, neither of the above two methods fully leveraged the complementary information inside RGB-D data. 
 For UR\&R, point clouds are only used for localization but do not participate in feature extraction. 
 For BYOC, their 3D descriptors are limited by their single-modality teacher. 
 To address above the problem, LLT \cite{ref14} introduced a linear transformer-based attention \cite{ref15,ref16} to embed geometric features into visual features in the last two stages. 
 This fusion improves extracted features and helps LLT achieve the state-of-the-art performance. 
 Whereas we believe unidirectional fusion in late stages does not fully exploit the complementary information in RGB-D data. 
 Therefore, we design a multi-scale bidirectional fusion network, which implements bidirectional fusion in all stages. 
 Benefiting from our fusion strategy, our network can achieve better performance with easily accessible backbones than unidirectional fusion with sophisticated backbones in LLT.

 \subsection{RGB-D Fusion}
 RGB image commonly contains rich semantic information, while depth image or point cloud can provide precise geometric description. 
 Therefore, fusing these two modalities is a promising direction as they provide complementary information. 
 With the advance of inexpensive RGB-D sensors, many works have studied how to fully leverage this complementary information in many tasks such as detection \cite{ref23,ref24,ref25,ref26,ref27,ref28,reviewer3}, segmentation \cite{ref29,ref30,ref31,ref32,ref33,ref34,reviewer1,reviewer2,reviewer5,reviewer6} and pose estimation \cite{ref35,ref36,ref37}. 
 As shown in Figure \ref{fig1}, the common fusion strategies can be roughly divided into three categories according to their information flow direction. 
 The first category is \textbf{undirected fusion} \cite{ref25,ref30,ref36,ref37,reviewer4}. 
 This category is the most intuitive one and is commonly implemented by directly concatenating or adding the separately extracted features. 
 For example, DenseFusion \cite{ref37} fuses geometric information and texture information by concatenating the embeddings from CNN and PointNet \cite{ref20} and adding extra channels for global information. 
 The second category is \textbf{unidirectional fusion} \cite{ref23,ref26,ref27,ref28,ref14,ref31,ref32,ref33,ref34,huang1,huang2,reviewer5}. 
 This kind of methods usually use one modality to guide the other modality. 
 For instance, DeepFusion \cite{ref23} sets Lidar features as queries and utilizes a cross-attention-based module called LearnableAlign to embed RGB image features into them. 
 % For instance, depth-aware CNN \cite{ref33} designs a deep-aware convolution to avoid ambiguity. 
 % DeepFusion \cite{ref23} set Lidar features as queries and utilizes a cross-attention-based module called LearnableAlign to embed camera features into them. 
 Similar to DeepFusion, LLT \cite{ref14} adopts a fusion module which is based on linear transformer \cite{ref15} and fuses high-level features in the last two layers. 
 However, all above methods do not fully exploit the interconnection between different modalities. 
 Therefore, the third category i.e. \textbf{bidirectional fusion} \cite{ref62,ref29,ref35,reviewer6} was proposed recently. 
 BPNet \cite{ref29} reveals that joint optimization on different modalities in a bidirectional manner is beneficial to 2D/3D semantic segmentation. 
 It designs a bidirectional projection module to generate a link matrix i.e. the point-pixel-wise map, so that information can interact between two heterogeneous network branches in the decoding stage. 
 FFB6D \cite{ref35} proposes a network fusing in full stages and outperforms previous methods \cite{ref36,ref37} a lot in pose estimation. 
 Motivated by these success, we believe bidirectional fusion can better leverage the complementary information inside two different domains and propose a bidirectional fusion-based network for RGB-D point cloud registration for the first time.

 % \section{Image to point cloud}
 % A image can be considered as a projection of a 3D object. 
 % For a calibrated RGB-D sensor, if the depth image is known, the 3D coordinate corresponding to each RGB pixel can be inferred by the inverse of projection:
 % \begin{equation}
 %   p_i=z_i K^{-1}\left[u_i, v_i, 1\right]^T
 % \end{equation}\label{eq1}
 % where $K$ denotes the intrinsic matrix of the sensor, and $p_i = [x_i, y_i, z_i]^T$ denotes the 3D point in the sensor coordinate system corresponding to the RGB pixel located at $[u_i, v_i]^T$. 
 % Since we use multi-scale point-pixel fusion in our work, it is necessary to generate point-pixel maps in multi-scales. 
 % In this work, we use a mean pooling operator to generate depth images in multi-scales and build maps between pixels and points in multi-scales by back-projecting pixels to 3D points. 
 
 \begin{figure*}[ht]
   \begin{center}
   \centerline{\includegraphics[width=2.0\columnwidth]{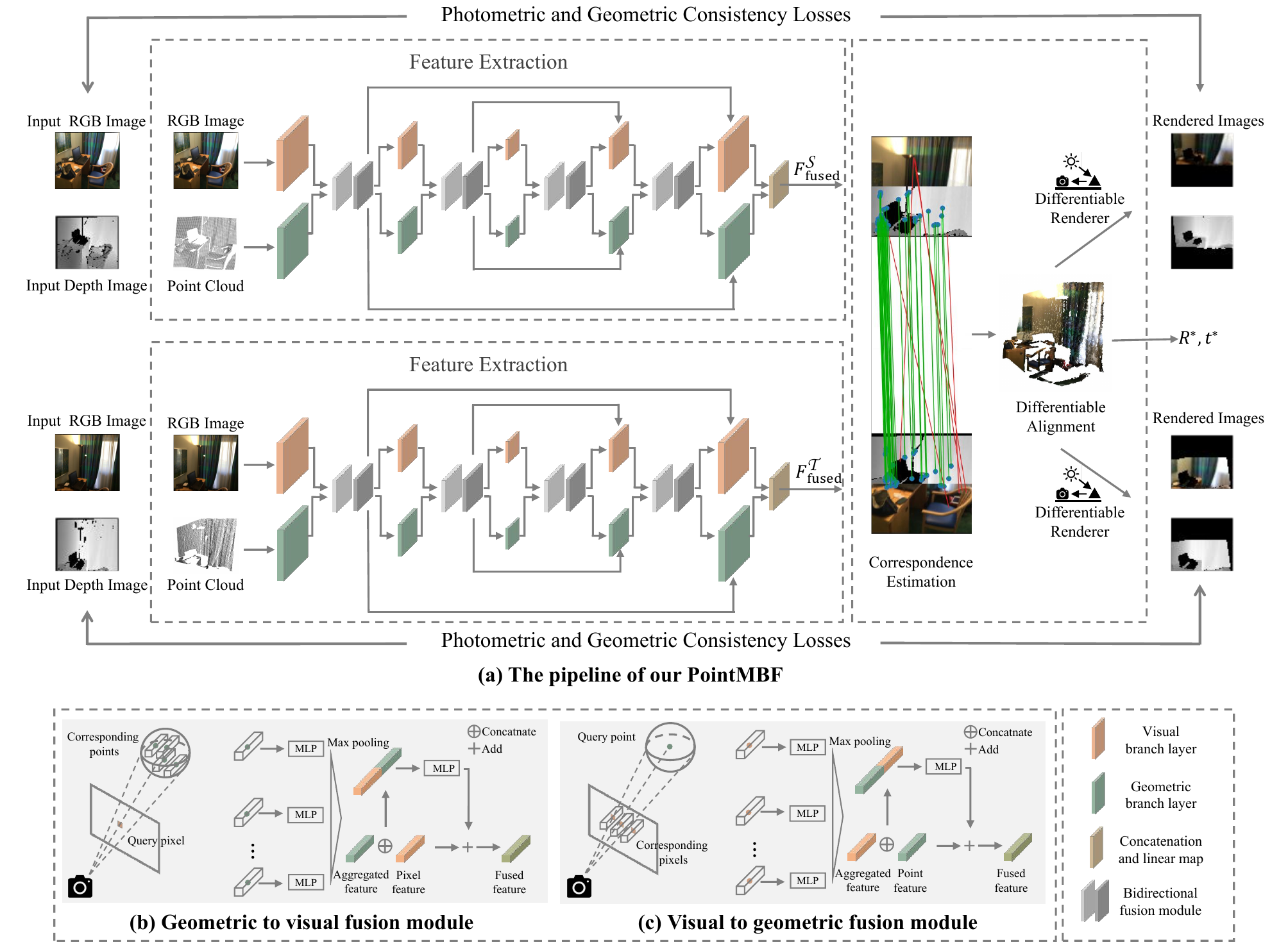}}
   \caption{
     \textbf{The overview of our PointMBF.} 
     It takes two RGB-D images as inputs and outputs an estimated rigid transformation. 
     For input RGB-D pairs, it first extracts features using a multi-scale bidirectional fusion-based extractor, which contains two branches and fusion modules (colored in grey) for feature interaction. 
     Then the putative correspondences are determined based on the Lowe's ratio of the extracted features. 
     Once obtaining the correspondences, our model outputs the estimated transformation using several RANSAC iterations. The above model is trained end-to-end by a differentiable renderer. 
     }
   \label{fig2}
   \end{center}
   % \vspace{-20pt}
 \end{figure*}

 \section{Method}
 Figure \ref{fig2} (a) shows the pipeline of our PointMBF, which takes two RGB-D images as inputs and outputs their relative rigid transformation represented by a rotation $R^*$ and a translation $t^*$. 
 Our PointMBF also follows the standard process of feature extraction, correspondence estimation and geometric fitting. 
 % Given two RGB-D images, our PointMBF first extracts deep features using two heterogeneous network branches for each of the two input RGB-D images, where the visual and geometric features are extracted using different networks and they are fully fused in a bidirectional manner in all stages by fusion modules. 
 PointMBF first extracts deep features using two heterogeneous network branches for each of the two input RGB-D images, where the visual and geometric features are extracted using different networks and they are fully fused in a bidirectional manner in all stages by fusion modules. 
 Then the fused features are used to generate correspondences based on their Lowe's ratio \cite{ref5}. 
 Finally, our PointMBF outputs the estimated rigid transformation using these correspondences by a few RANSAC iterations. 
 The correspondence estimation and geometric fitting are free of learnable parameters, and our feature extractor and the fusion modules are trained unsupervisedly by a renderer-based loss. 
 The details of each component of our PointMBF are explained in the following sections. 
 
 \subsection{Heterogeneous Network Branches}\label{sec31}
 Since there exists a big domain gap between RGB images and depth images, our PointMBF uses two different network branches to process these two modalities separately. 
 As shown in Figure \ref{fig2} (a), one branch i.e the visual branch takes RGB images as input, while the other i.e. the geometric branch takes point clouds generated from depth images as input. 
 Both branches follows a U-Shape \cite{ref18} structure to extract multi-scale information, and they are all based on easily accessible backbones including ResNet18 \cite{ref19} and KPFCN \cite{ref41,ref17}. 
 Since our competitor LLT \cite{ref14} also has two branches for visual and geometric processing, we introduce the details of our two branches in the following paragraphs and compare them with similar structures in LLT. 
 
 \noindent
 \textbf{Visual branch.} 
 LLT designs a dilated convolution-based network as its visual backbone. 
 Although this kind of backbone is competitive, its performance is highly dependent on the hyperparameter setting. 
 To better illustrate the effectiveness of our multi-scale bidirectional fusion strategy and save cost on tuning the network architecture, we simply modify a widely used ResNet18 \cite{ref19} as our visual branch. 
 
 As shown in Figure \ref{fig2} (a), our visual branch follows an U-Shape encoder-decoder architecture with skip connections. 
 Both encoder and decoder extracts features at three different scales. 
 The encoder consists of convolution blocks from ResNet18, while the decoder only contains simple shallow convolution blocks. 
 More details of our visual branch settings are provided in the supplementary materials.

 \noindent
 \textbf{Geometric branch.} 
 Different from LLT \cite{ref14}, we process depth images in the form of point clouds rather than the original depth images. 
 There exist many feature extractors for point cloud such as sparse convolution networks \cite{ref3,ref51}, point-based networks \cite{ref20,ref60} and so on. 
 However, as shown in Figure \ref{fig3}, there exists severe density variation in the generated point clouds because the sampling density of 3D surfaces is dependent on their distance to the sensor. 
 To extract density-invariant features, we select a shallower KPFCN in D3Feat \cite{ref41} as the building block of our geometric branch because it introduces a density normalization process to overcome the inherent density variation. 
 
 As shown in Figure \ref{fig2} (a), our geometric branch has a symmetric architecture to the visual branch, so that features from the two branches at the same resolution can be fused and this kind of fusion occurs at every scales. 
 More details of our geometric branch settings are also provided in the supplementary materials.

 \subsection{Multi-scale Bidirectional Fusion}
 In this section, we introduce our proposed multi-scale bidirectional fusion in detail. Note that semantics or local geometry are dependent on a certain region rather than a single pixel or point. 
 Therefore, it is intuitive to fuse complementary information by embedding features of a certain region into features of the other modality. 
 
\begin{figure}[ht]
   \begin{center}
   \centerline{\includegraphics[width=0.8\columnwidth]{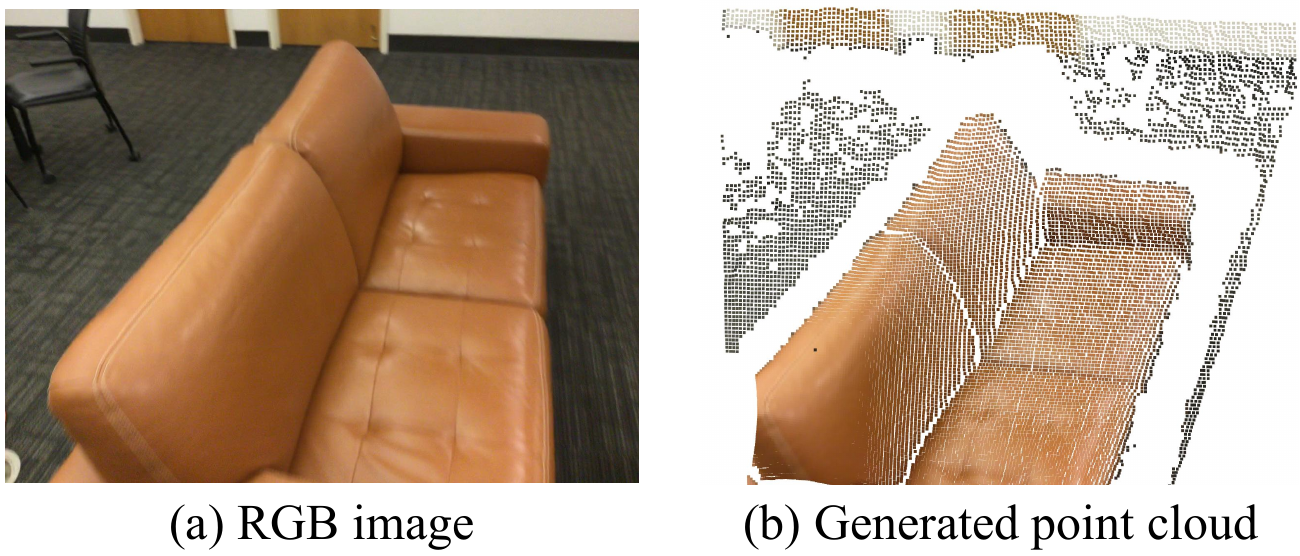}}
   \caption{
     Input RGB image (a) and the point cloud (slightly rotated) generated from the corresponding depth image (b). 
     Severe density variation exists in the generated point cloud, which makes local geometric feature extraction and fusion more challenging. 
     }
   \label{fig3}
   \end{center}
   % \vspace{-20pt}
 \end{figure}

 However, embedding regional features faces two challenges. 
 First, as shown in Figure \ref{fig3}, density variation makes the length of regional feature set uncertain. 
 Second, the feature set is not structural data. 
 Inspired by the process for variable length sequence \cite{ref63} and unstructured data \cite{ref20}, we pad the regional features to a fixed number and design a PointNet-style \cite{ref20} fusion module for bidirectional fusion. 
 As shown in Figure \ref{fig2} (b)(c), for a query pixel or a query 3D point at a certain scale, we first find its corresponding region in point cloud/image using the intrinsic matrix of the sensor. 
 Afterward, we sample the KNN corresponding points/pixels in the corresponding region and gather their features to a set. 
 The set is then padded to a certain length and aggregated by a simple PointNet. 
 Since there exists a max-pooling operator in PointNet and grid sampling in the geometric branch, the aggregated feature can achieve density-invariance. 
 Finally, the aggregated feature is further fused with the feature of the query point/pixel by a shallow neural network with residue design. 
 In this way, visual and geometric features can be fully fused in all scales. 
 Besides, as shown in Figure \ref{fig2} (a), in addition to the above fusion using bidirectional fusion module, we also conduct an undirected fusion in the final stage to further boost the features for correspondence estimation. 
 Details of the visual-to-geometric, the geometric-to-visual, and the final undirected fusion will be introduced in the following subsections.

 \noindent
 \textbf{Visual-to-geometric fusion.} 
 Commonly, many ambiguous and repetitive structures exist in point clouds, which makes generated putative correspondences based on only point clouds contain a large proportion of outliers. 
 Incorporating semantic information extracted by the visual branch can make geometric features more distinctive. 
 Here, we utilize visual-to-geometric fusion to embed regional visual features into geometric features.
 
 Specifically, given a geometric feature $F^l_{g_i}$ extracted by the geometric branch in the $l$-th stage for the $i$-th point, we first find its corresponding region in the image by projecting its neighbor with radius $R^l_{v2g}$ to the image. 
 Then we sample the $K_{v2g}$ nearest neighbor pixels within the corresponding region and gather their visual features $\left\{F_{v_k}^l\right\}_{k=1}^{K_{v 2 g}}$. 
 If there are less than $K_{v2g}$ pixels in the corresponding region, we will pad the null feature $F_{p a d}=[0,0, \ldots, 0] \in \mathbb{R}^{d^l}$ in the gathered features. 
 After that, we use a PointNet-style fusion module to aggregate the regional visual features $\left\{F_{v_k}^l\right\}_{k=1}^{K_{v 2 g}}$:
 \begin{equation}
   F_{v 2 g_i}^l=\max _{k=1}^{K_{v 2 g}}\left(\operatorname{MLP}\left(F_{v_k}^l\right)\right)
 \end{equation}

 We then concatenate the aggregated feature $F_{v 2 g_i}^l$ with the geometric feature $F_{g_i}^l$ and use a linear layer to map them to a fused feature, which has the same dimension as $F_{g_i}^l$. 
 Finally, this fused feature is treated as a residue and added to the original geometric feature $F_{g_i}^l$: 
 \begin{equation}
   F_{\text {fused}g_i}^l=F_{g_i}^l+W_{v 2 g}^l\left(F_{v 2 g_i}^l \oplus F_{g_i}^l\right)
 \end{equation}
 where $\oplus$ denotes the concatenate operation, $W_{v 2 g}^l$ denotes a linear map, and $F_{\text {fused}g_i}^l$ denotes the final fused feature, which replaces the original geometric feature $F_{g_i}^l$ to be sent to the next stage. 
 In our bidirectional fusion modules, we adopt a residual design, since our full-stage fusion may cause redundancy. 
 Our following ablation study also verifies it experimentally.

 \noindent
 \textbf{Geometric-to-visual fusion.} 
 Similar to visual-to-geometric fusion, our geometric-to-visual fusion also makes visual features more distinctive. 
 We achieve geometric-to-visual fusion by embedding geometric features into visual features. 
 
 Given a visual feature $F_{v_i}^l$ extracted by visual branch in $l$-th stage for the $i$-th pixel, we first find its corresponding region in the 3D point clouds by the inverse projection. 
 Then we sample $K_{g2v}$ nearest neighbor points in the corresponding region and gather their geometric features $\left\{F_{g_k}^l\right\}_{k=1}^{K_{g 2 v}}$. 
 We also use the null feature $F_{p a d}=[0,0, \ldots, 0] \in \mathbb{R}^{d^l}$ to pad the gathered features when there are not enough points in the corresponding region and aggregate them into $F_{g2v_i}^l$: 
 \begin{equation}
   F_{g 2 v_i}^l=\max _{k=1}^{K_{g 2 v}}\left(\operatorname{MLP}\left(F_{g_k}^l\right)\right)
 \end{equation}
 
 The aggregated feature $F_{g 2 v_i}^l$ is concatenated with the visual feature $F_{v_i}^l$ and then mapped to a feature, which has the same dimension as $F_{v_i}^l$. 
 Finally, this fused feature is treated as a a residue and added to the original visual feature $F_{v_i}^l$: 
 \begin{equation}
   F_{\text {fused}v_i}^l=F_{v_i}^l+W_{g 2 v}^l\left(F_{g 2 v_i}^l \oplus F_{v_i}^l\right)
 \end{equation}
 where $\oplus$ denotes the concatenate operation, $W_{g 2 v}^l$ denotes a linear map, and $F_{\text {fused}v_i}^l$ denotes the final fused feature, which replace the original visual feature $F_{v_i}^l$ to be sent to the next stage.

 \noindent
 \textbf{Undirected fusion.} 
 After fully bidirectional fusion in both encoding and decoding stages, we have obtained distinctive features extracted by the visual and geometric branches. 
 To obtain more distinctive features for generating reliable correspondences, we use a simple undirected fusion in the final stage. 
 We concatenate the outputs of both the visual and the geometric branches and fuse them by a linear map:
 \begin{equation}
   F_{\text{fused}_i}=W_{\text {final }}\left(F_{g_i}^{\text {final}} \oplus F_{v_i}^{\text {final }}\right)
 \end{equation}
 where $W_{\text {final }}$ denotes a linear map, $F_{g_i}^{\text {final}}$ denotes the geometric feature output by the last layer of geometric branch, $F_{v_i}^{\text {final}}$ denotes the visual feature output by the last layer of visual branch, and $F_{\text{fused}_i}$ denotes the final fused features for the following correspondence estimation. 
 % These features are fed into following correspondence estimation module
 
 \subsection{Correspondence Estimation, Geometric Fitting and Loss Function}
 
 \noindent
 \textbf{Correspondence estimation and geometric fitting.} 
 After obtaining the fused features for the source and the target point clouds, we build correspondences using the same method as in UR\&R \cite{ref12} and LLT \cite{ref14}. 
 Specifically, the correspondences are generated based on the the lowe's ratio \cite{ref5}. 
 For a point $p_i^\mathcal{S}$ in the source point cloud, the lowe's ratio $r_i^\mathcal{S}$ is formulated as:
 \begin{equation}
   r_i^\mathcal{S}=\frac{D\left(p_i^\mathcal{S}, p_{1 n n}^\mathcal{T}\right)}{D\left(p_i^\mathcal{S}, p_{2 n n}^\mathcal{T}\right)}
 \end{equation}
 where $D(\cdot)$ denotes the Euclidean distance in the feature space and $p_{knn}^\mathcal{T}$ is the $k$-th similar point in the target point cloud. 
 Then we calculate the weight $w=1-r$ for each correspondence and select the correspondences with top $k$ weights for source point cloud and target point cloud respectively. 
 The selected correspondences $C=\left\{\left(p^\mathcal{S}, p^\mathcal{T}, w\right)_i: 0 \leq i<2 k\right\}$ with their weights are fed into a RANSAC \cite{ref6} module. 
 % The RANSAC module outputs an estimated rigid transformation $T^*$ with the minimum error $E(C,T^*)$, where $E(C,T)$ is formulated as:
 The RANSAC module achieves differentiable alignment and outputs an estimated rigid transformation $T^*$ with the minimum error $E(C,T^*)$, where $E(C,T)$ is formulated as:
 \begin{equation}
   E(C, T)=\sum_{\left(p^\mathcal{S}, p^\mathcal{T}, w\right) \in C} w\left(p^\mathcal{S}-T\left(p^\mathcal{T}\right)\right)^2 / 2 k
   \label{eq7}
  \end{equation}
 
 \noindent
 \textbf{Loss function.}
 % In this work, we use the same loss function as UR\&R \cite{ref12} and LLT \cite{ref14} to train the model without the need for annotation. 
 In this work, we use the same loss function as \cite{ref12,ref14} to train the model without the need for annotation. 
 The loss function consists three components:
 \begin{equation}
   \mathcal{L}=l_{g e o}+l_{v i s}+\lambda E\left(C, T^*\right)
 \end{equation}
 where $l_{g e o}$ and $l_{v i s}$ denote the geometric and photometric losses based on a differentiable renderer, $\lambda$ represents a coefficient and we set $\lambda=0.1$. 
 More details about the loss function can be found in UR\&R \cite{ref12}.

 \begin{table*}[ht]
   \centering
   \caption{\textbf{Pairwise registration on ScanNet \cite{ref21}.} 
   Pose Sup indicates the pose or correspondence supervision.}
   \scalebox{0.75}{
   \begin{tabular}{lccccccccccccccccc}
   \hline
                       &                             &                            & \multicolumn{5}{c}{Rotation (deg)}                                          & \multicolumn{5}{c}{Translation (cm)}                                        & \multicolumn{5}{c}{Chamfer (mm)}                                            \\
                       &                             &                            & \multicolumn{3}{c}{Accuracy↑}                 & \multicolumn{2}{c}{Error↓}  & \multicolumn{3}{c}{Accuracy↑}                 & \multicolumn{2}{c}{Error↓}  & \multicolumn{3}{c}{Accuracy↑}                 & \multicolumn{2}{c}{Error↓}  \\ \cline{4-18} 
   \multirow{-3}{*}{}  & \multirow{-3}{*}{Train Set} & \multirow{-3}{*}{Pose Sup} & 5             & 10            & 45            & Mean         & Med.         & 5             & 10            & 25            & Mean         & Med.         & 1             & 5             & 10            & Mean         & Med.         \\ \hline
   ICP \cite{ref42}        & -                           &                            & 31.7          & 55.6          & \textbf{99.6} & 10.4         & 8.8          & 7.5           & 19.4          & 74.6          & 22.4         & 20.0         & 8.4           & 24.7          & 40.5          & 32.9         & 14.1         \\
   FPFH \cite{ref2}        & -                           &                            & 34.1          & 64.0          & 90.3          & 20.6         & 7.2          & 8.8           & 26.7          & 66.8          & 42.6         & 18.6         & 27.0          & 60.8          & 73.3          & 23.3         & 2.9          \\
   SIFT \cite{ref5}        & -                           &                            & 55.2          & 75.7          & 89.2          & 18.6         & 4.3          & 17.7          & 44.5          & 79.8          & 26.5         & 11.2         & 38.1          & 70.6          & 78.3          & 42.6         & 1.7          \\
   SuperPoint \cite{ref56} & -                           &                            & 65.5          & 86.9          & 96.6          & 8.9          & 3.6          & 21.2          & 51.7          & 88.0          & 16.1         & 9.7          & 45.7          & 81.1          & 88.2          & 19.2         & 1.2          \\
   FCGF \cite{ref3}        & -                           & $\checkmark$               & 70.2          & 87.7          & 96.2          & 9.5          & 3.3          & 27.5          & 58.3          & 82.9          & 23.6         & 8.3          & 52.0          & 78.0          & 83.7          & 24.4         & 0.9          \\ \hline
   DGR \cite{ref39}       & 3DMatch                     & $\checkmark$               & 81.1          & 89.3          & 94.8          & 9.4          & 1.8          & 54.5          & 76.2          & 88.7          & 18.4         & 4.5          & 70.5          & 85.5          & 89.0          & 13.7         & 0.4          \\
   3D MV Reg \cite{ref57}  & 3DMatch                     & $\checkmark$              & 87.7          & 93.2          & 97.0          & 6.0          & 1.2          & 69.0          & 83.1          & 91.8          & 11.7         & 2.9          & 78.9          & 89.2          & 91.8          & 10.2         & 0.2          \\
   REGTR \cite{ref38}     & 3DMatch                     & $\checkmark$               & 86.0          & 93,9          & 98.6          & 4.4          & 1.6          & 61.4          & 80.3          & 91.4          & 14.4         & 3.8          & 80.9          & 90.9          & 93.6          & 13.5         & 0.2          \\
   UR\&R \cite{ref12}      & 3DMatch                     &                            & 87.6          & 93.1          & 98.3          & 4.3          & 1.0          & 69.2          & 84.0          & 93.8          & 9.5          & 2.8          & 79.7          & 91.3          & 94.0          & 7.2          & 0.2          \\
   UR\&R (RGB-D)       & 3DMatch                     &                            & 87.6          & 93.7          & 98.8          & 3.8          & 1.1          & 67.5          & 83.8          & 94.6          & 8.5          & 3.0          & 78.6          & 91.7          & 94.6          & 6.5          & 0.2          \\
   UR\&R (Supervised)  & 3DMatch                     & $\checkmark$                & 92.3          & 95.3          & 98.2          & 3.8          & \textbf{0.8} & 77.6          & 89.4          & 95.5          & 7.8          & 2.3          & 86.1          & 94.0          & 95.6          & 6.7          & \textbf{0.1} \\
   BYOC \cite{ref13}       & 3DMatch                     &                            & 66.5          & 85.2          & 97.8          & 7.4          & 3.3          & 30.7          & 57.6          & 88.9          & 16.0         & 8.2          & 54.1          & 82.8          & 89.5          & 9.5          & 0.9          \\
   LLT \cite{ref14}       & 3DMatch                     &                            & 93.4          & 96.5          & 98.8          & \textbf{2.5} & \textbf{0.8} & 76.9          & 90.2          & 96.7          & \textbf{5.5} & 2.2          & 86.4          & 95.1          & 96.8          & \textbf{4.6} & \textbf{0.1} \\
   \rowcolor[HTML]{EFEFEF} 
   Ours                & 3DMatch                     &                            & \textbf{94.6} & \textbf{97.0} & 98.7          & 3.0          & \textbf{0.8} & \textbf{81.0} & \textbf{92.0} & \textbf{97.1} & 6.2          & \textbf{2.1} & \textbf{91.3} & \textbf{96.6} & \textbf{97.4} & 4.9          & \textbf{0.1} \\ \hline
   UR\&R \cite{ref12}      & ScanNet                     &                            & 92.7          & 95.8          & 98.5          & 3.4          & 0.8          & 77.2          & 89.6          & 96.1          & 7.3          & 2.3          & 86.0          & 94.6          & 96.1          & 5.9          & \textbf{0.1} \\
   UR\&R (RGB-D)       & ScanNet                     &                            & 94.1          & 97.0          & 99.1          & 2.6          & 0.8          & 78.4          & 91.1          & 97.3          & 5.9          & 2.3          & 87.3          & 95.6          & 97.2          & 5.0          & \textbf{0.1} \\
   BYOC \cite{ref13}       & ScanNet                     &                            & 86.5          & 95.2          & 99.1          & 3.8          & 1.7          & 56.4          & 80.6          & 96.3          & 8.7          & 4.3          & 78.1          & 93.9          & 96.4          & 5.6          & 0.3          \\
   LLT \cite{ref14}       & ScanNet                     &                            & 95.5          & \textbf{97.6} & 99.1          & \textbf{2.5} & 0.8          & 80.4          & 92.2          & 97.6          & \textbf{5.5} & 2.2          & 88.9          & 96.4          & 97.6          & \textbf{4.6} & \textbf{0.1} \\
   \rowcolor[HTML]{EFEFEF} 
   Ours                & ScanNet                     &                            & \textbf{96.0} & \textbf{97.6} & 98.9          & \textbf{2.5} & \textbf{0.7} & \textbf{83.9} & \textbf{93.8} & \textbf{97.7} & 5.6          & \textbf{1.9} & \textbf{92.8} & \textbf{97.3} & \textbf{97.9} & 4.7          & \textbf{0.1} \\ \hline
   \end{tabular}
   }
   \label{table1}
 \end{table*}

 \begin{table*}[]
   \centering
   \caption{
    % \textbf{Comparison with other fusion strategies and single branch performance of our method and LLT \cite{ref14}.} 
    \textbf{Single branch performance of our method and LLT \cite{ref14} (upper rows) and comparison with other fusion strategies (lower rows).} 
  %  Visual denotes the visual branch, Geo denotes the geometric branch. 
   Visual (Ours) and Geo (Ours) denote the visual and geometric branches of our PointMBF, respectively.
   Visual (LLT) denotes the visual branch in LLT, which is based on the dilated convolution. 
   Visual (RGB-D) denotes our visual branch with an additional channel for depth images. 
   All these networks can resemble augmented version of UR\&R \cite{ref12} with different feature extractors. 
   CAT denotes fusion using direct concatenation. 
   DF denotes fusion using DenseFusion \cite{ref37}. 
   Trans denotes fusion using transformer \cite{ref16,ref23} in high-level feature space like DeepFusion \cite{ref23}. 
   Ours wo res denotes removing the residue design in our fusion modules. }
   \scalebox{0.75}{
     \setlength{\tabcolsep}{3mm}{
   \begin{tabular}{lccccccccccccccc}
   \hline
                         & \multicolumn{5}{c}{Rotation (deg)}                                          & \multicolumn{5}{c}{Translation (cm)}                                        & \multicolumn{5}{c}{Chamfer (mm)}                                            \\
                         & \multicolumn{3}{c}{Accuracy↑}                 & \multicolumn{2}{c}{Error↓}  & \multicolumn{3}{c}{Accuracy↑}                 & \multicolumn{2}{c}{Error↓}  & \multicolumn{3}{c}{Accuracy↑}                 & \multicolumn{2}{c}{Error↓}  \\ \cline{2-16} 
                         & 5            & 10           & 45          & Mean         & Med.         & 5             & 10            & 25            & Mean         & Med.         & 1             & 5             & 10            & Mean         & Med.         \\ \hline
                         Visual (Ours)                & 89.9          & 94.3          & 98.4          & 3.9          & 1.0          & 72.4          & 86.7          & 94.9          & 8.4          & 2.6          & 82.7          & 92.8          & 95.1          & 6.7          & 0.2          \\
                         Geo (Ours)                  & 32.8          & 61.9          & 93.4          & 15.9         & 7.5          & 11.7          & 27.6          & 62.1          & 36.1         & 18.5         & 24.1          & 54.0          & 67.7          & 21.8         & 4.1          \\
   Visual (LLT \cite{ref14}) & 90.4          & 95.0          & 98.6          & 3.6          & 1.0          & 70.8          & 86.5          & 95.3          & 8.1          & 2.8          & 81.8          & 93.1          & 95.4          & 6.2          & 0.2          \\ \hline
   Visual (RGB-D)         & 85.0          & 92.1          & 98.2          & 4.7          & 1.1          & 64.1          & 80.6          & 92.7          & 10.2         & 3.3          & 75.8          & 89.3          & 92.8          & 7.7          & 0.2          \\ 
   CAT                   & 93.1          & 96.1          & \textbf{98.7} & 3.2          & \textbf{0.8} & 78.5          & 90.5          & 96.4          & 6.7          & 2.2          & 89.7          & 95.7          & 96.9          & 5.6          & \textbf{0.1} \\
   DF \cite{ref37}           & 92.9          & 96.0          & 98.6          & 3.3          & \textbf{0.8} & 78.2          & 90.3          & 96.3          & 6.8          & 2.2          & 89.2          & 95.6          & 96.8          & 5.4          & \textbf{0.1} \\
   Trans \cite{ref23}       & 91.5          & 95.2          & 98.3          & 3.6          & 0.9          & 74.7          & 88.1          & 95.6          & 7.7          & 2.5          & 87.3          & 94.8          & 96.3          & 5.6          & \textbf{0.1} \\
   \rowcolor[HTML]{EFEFEF} 
   Ours wo res           & 94.0          & 96.6          & \textbf{98.7} & 3.1          & \textbf{0.8} & 80.3          & 91.3          & 96.8          & 6.3          & \textbf{2.1} & 90.7          & 96.2          & 97.2          & 5.3          & \textbf{0.1} \\
   \rowcolor[HTML]{EFEFEF} 
   Ours                  & \textbf{94.6} & \textbf{97.0} & \textbf{98.7} & \textbf{3.0} & \textbf{0.8} & \textbf{81.0} & \textbf{92.0} & \textbf{97.1} & \textbf{6.2} & \textbf{2.1} & \textbf{91.3} & \textbf{96.6} & \textbf{97.4} & \textbf{4.9} & \textbf{0.1} \\ \hline
   \end{tabular}
   }
   }
   \label{table2}
 \end{table*}
 
 \begin{table*}[]
   \centering
   \caption{\textbf{Ablation on fusion stages.} 
   Encode denotes bidirectional fusion during encoding stage. 
   Decode denotes bidirectional fusion during decoding stage. 
   Concat denotes the concatenation and linear map in the final stage. }
   \scalebox{0.75}{
     \setlength{\tabcolsep}{2.6mm}{
   \begin{tabular}{cccccccccccccccccc}
   \hline
   Encode & Decode & Concat & \multicolumn{5}{c}{Rotation (deg)}                                                & \multicolumn{5}{c}{Translation (cm)}                                             & \multicolumn{5}{c}{Chamfer (mm)}                                                 \\ 
        &    &     & \multicolumn{3}{c}{Accuracy↑}                 & \multicolumn{2}{c}{Error↓}  & \multicolumn{3}{c}{Accuracy↑}                 & \multicolumn{2}{c}{Error↓}  & \multicolumn{3}{c}{Accuracy↑}                 & \multicolumn{2}{c}{Error↓}  \\ \cline{4-18} 
        &    &     & 5            & 10           & 45           & Mean         & Med.         & 5             & 10            & 25            & Mean         & Med.         & 1             & 5             & 10            & Mean         & Med.         \\ \hline
        $\checkmark$  &    &     & 94.1          & 96.7          & \textbf{98.7} & 3.1          & \textbf{0.8} & 80.1          & 91.5          & 96.8          & 6.4          & \textbf{2.1} & 90.7          & 96.3          & 97.3          & 5.5          & \textbf{0.1} \\
        & $\checkmark$   &     & 93.9          & 96.6          & 98.5          & 3.2          & 0.8          & 79.8          & 91.5          & 96.8          & 6.7          & 2.2          & 90.6          & 96.2          & 97.2          & 5.4          & \textbf{0.1} \\
        &    & $\checkmark$    & 93.1          & 96.1          & \textbf{98.7} & 3.2          & \textbf{0.8} & 78.5          & 90.5          & 96.4          & 6.7          & 2.2          & 89.7          & 95.7          & 96.9          & 5.6          & \textbf{0.1} \\
        $\checkmark$   &    & $\checkmark$    & 94.3          & 96.8          & \textbf{98.7} & \textbf{3.0} & \textbf{0.8} & 80.5          & \textbf{92.0} & 97.0          & \textbf{6.1} & \textbf{2.1} & 91.0          & 96.4          & \textbf{97.4} & 5.1          & \textbf{0.1} \\
        & $\checkmark$   & $\checkmark$    & 94.1          & 96.5          & 98.6          & 3.2          & \textbf{0.8} & 80.6          & 91.6          & 96.7          & 6.5          & \textbf{2.1} & 91.0          & 96.2          & 97.2          & 5.5          & \textbf{0.1} \\
        $\checkmark$   & $\checkmark$   &     & 94.2          & 96.7          & \textbf{98.7} & 3.1          & \textbf{0.8} & 80.3          & 91.7          & 96.9          & 6.3          & 2.2          & 90.8          & 96.3          & 97.3          & 5.4          & \textbf{0.1} \\
        $\checkmark$   & $\checkmark$   & $\checkmark$   & \textbf{94.6} & \textbf{97.0} & \textbf{98.7} & \textbf{3.0} & \textbf{0.8} & \textbf{81.0} & \textbf{92.0} & \textbf{97.1} & 6.2          & \textbf{2.1} & \textbf{91.3} & \textbf{96.6} & \textbf{97.4} & \textbf{4.9} & \textbf{0.1} \\ \hline
     \end{tabular}
     }
   }
   \label{table3}
   \end{table*}

   \begin{table*}[]
     \centering
   \caption{\textbf{Ablation on fusion direction.} 
   V2G denotes reserving fusion from the visual branch to the geometric branch. 
   G2V denotes reserving fusion from the geometric branch to the visual branch. 
   CAT denotes reserving final undirected fusion i.e. concatenation. 
   }
   \scalebox{0.75}{
     \setlength{\tabcolsep}{2.8mm}{
     \begin{tabular}{cccccccccccccccccc}
     \hline
     V2G & G2V & CAT & \multicolumn{5}{c}{Rotation (deg)}                                                & \multicolumn{5}{c}{Translation (cm)}                                             & \multicolumn{5}{c}{Chamfer (mm)}                                                 \\
     &   &  & \multicolumn{3}{c}{Accuracy↑}                 & \multicolumn{2}{c}{Error↓}  & \multicolumn{3}{c}{Accuracy↑}                 & \multicolumn{2}{c}{Error↓}  & \multicolumn{3}{c}{Accuracy↑}                 & \multicolumn{2}{c}{Error↓}  \\ \cline{4-18} 
     &   &  & 5            & 10           & 45           & Mean         & Med.         & 5             & 10            & 25            & Mean         & Med.         & 1             & 5             & 10            & Mean         & Med.         \\ \hline
         $\checkmark$   &   &  &91.4&  95.8 &    98.4&  3.8&    1.1 &  72.6&    88.3 &    96.0 &   7.8 & 2.8 &    87.1 &    95.2 &   96.5&    5.9 &   \textbf{0.1}   \\
               & $\checkmark$  &  &   93.1    &  96.1  & 98.5 &  3.4  & \textbf{0.8} &  78.1  &  90.4  &  96.4 & 7.0  &  2.3   &  89.5&   95.9 & 96.8  & 5.5 & \textbf{0.1} \\
         &  &  $\checkmark$ & 93.1          & 96.1          & \textbf{98.7} & 3.2          & \textbf{0.8} & 78.5          & 90.5          & 96.4          & 6.7          & 2.2          & 89.7          & 95.7          & 96.9          & 5.6          & \textbf{0.1} \\
         $\checkmark$   &   &$\checkmark$   & 94.1          & 96.7          & \textbf{98.7} & 3.1          & \textbf{0.8} & 80.3          & 91.7          & 96.9          & 6.3          & \textbf{2.1} & 91.0          & 96.4          & 97.3          & 5.2          & \textbf{0.1} \\
          & $\checkmark$   & $\checkmark$ & 93.4          & 96.4          & \textbf{98.7} & 3.1          & \textbf{0.8} & 79.2          & 90.9          & 96.7          & 6.4          & \textbf{2.1} & 89.9          & 96.1          & 97.1          & 5.1          & \textbf{0.1} \\
          $\checkmark$   & $\checkmark$   & $\checkmark$  & \textbf{94.6} & \textbf{97.0} & \textbf{98.7} & \textbf{3.0} & \textbf{0.8} & \textbf{81.0} & \textbf{92.0} & \textbf{97.1} & \textbf{6.2} & \textbf{2.1} & \textbf{91.3} & \textbf{96.6} & \textbf{97.4} & \textbf{4.9} & \textbf{0.1} \\ \hline
     \end{tabular}
     }
     }
     \label{table4}
   %   \vspace{-0.5em}
     \end{table*}

 \section{Experiment}
 We follow the setting in UR\&R \cite{ref12} and use two indoor RGB-D datasets 3DMatch \cite{ref22} and ScanNet \cite{ref21} to conduct our experiments. 
 The following sections are organized as follows. 
 First, we illustrate the details of our experimental settings including datasets, implementation, evaluation metrics, and competitors in section \ref{sec51}. 
 Next, we evaluate our method on ScanNet in section \ref{sec52}. 
 In this section, we conduct two experiments. 
 The former tests the performance of our method trained on ScanNet \cite{ref21} and the latter tests our method trained on 3DMatch \cite{ref22} to verify its generalization. 
 To further understand the effect of our multi-scale bidirectional fusion, we conduct comprehensive ablation studies in section \ref{sec53}. 
 We also provide more visualizations and extra experiments in the supplementary material.

 \subsection{Experimental Settings}\label{sec51}
 
 \noindent
 \textbf{Datasets.} 
 We use two widely-used RGB-D datasets ScanNet \cite{ref21} and 3DMatch \cite{ref22}, which contain RGB-D images, camera intrinsics, and ground-truth poses of the camera. 
 % ScanNet contains RGB-D images, camera intrinsics, and ground-truth poses of camera of 1513 scenes. 
 % We follow the official split of 1045/156/312 scenes for train/val/test. 
 % 3DMatch contains 62 scenes. 
 % We follow \cite{ref12,ref13,ref14} to split it into 46/8/8 scenes for train/val/test. 
 For both datasets, we follow settings in \cite{ref12,ref13,ref14} to generate view pairs by sampling image pairs which are 20 frames apart. 
 This results in 1594k/12.6k/26k RGB-D pairs for ScanNet and 122k/1.5k/1.5k RGB-D pairs for 3DMatch for train/val/test, respectively. 
 
 \noindent
 \textbf{Implementation.} 
 To achieve a fair comparison, we use the same settings as LLT \cite{ref14} including batch size, learning rate, image size, and so on. 
 We set $K_{v2g} = 16$ for training and $K_{v2g} = 32$ for test. 
 Since pixels are more dense than valid points, we set $K_{g2v} = 1$ to save memory. 
 Before generating point clouds from depth images, we apply the hole completion algorithm \cite{ref61} to the depth images. 
 Our network is implemented in Pytorch \cite{ref55} and Pytorch3d \cite{ref53}. 
 All the experiments are conducted on a single A40 graphic card. 
 For more details of implementation, please see the supplementary material. 
 % More details of implementation can be found in the supplementary material.

 \noindent
 \textbf{Evaluation metrics.} 
 Following prior work \cite{ref12,ref13,ref14}, we evaluate the RGB-D point cloud registration by three evaluation metrics: rotation error, translation error, and chamfer error \cite{ref54}. 
 % The errors of rotation and translation are defined:
 % \begin{equation}
 %   \begin{gathered}
 %   E_{\text {rotation }}=\arccos \left(\frac{\operatorname{Tr}\left(R_{e s t}^T R_{g t}\right)-1}{2}\right) \\
 %   E_{\text {translation }}=\left\|t_{e s t}-t_{g t}\right\|_2
 %   \end{gathered}
 % \end{equation}
 % where $R_{e s t}$ and $t_{e s t}$ denote the estimation, $R_{gt}$ and $t_{gt}$ denote the ground-truth transformation. 
 % Although errors of rotation and translation can directly measure the registration performance, they are not a perfect measure for ambiguous scenes, especially when the scenes contain walls, floors, and symmetrical objects. 
 % Therefore, we also report the chamfer error of registered point clouds. 
 % Specifically, we define $P^{est}$ as a registered scene using the estimated transformation and define $P^{gt}$ as a registered scene using the ground-truth transformation. 
 % The chamfer error is defined as follows:
 % \begin{equation}
 %   E_{\text {chamfer }}=|P^{est}|^{-1}\sum_{(p^{{est}}, P^{gt}) \in \Lambda_{P^{est},P^{g t}}}||P^{est} - P^{gt}||_2 + 
 %   |P^{gt}|^{-1}\sum_{(p^{{gt}}, P^{est}) \in \Lambda_{P^{gt},P^{est}}}||P^{gt} - P^{est}||_2
 % \end{equation}
 % where $\Lambda_{P^{est},P^{gt}} = \{(p^{e s t}, \underset{p^{g t} \in P^{g t}}{\arg \min }||p^{est}-p^{g t}||_2), p^{est} \in P^{est}\}$ denotes a closest pairs between $P_{est}$ and $P_{gt}$, and vice versa for $\Lambda_{P^{gt},P^{est}}$. 
 For the above metrics, we not only report their mean and median values but also their accuracy under different thresholds. 
 
 \noindent
 \textbf{Competitors.} 
 Our competitors can be divided into three categories based on the modalities they use. The first category is only based on point cloud. 
 In addition to previous baselines including ICP \cite{ref42}, FPFH \cite{ref2}, FCGF \cite{ref3}, DGR \cite{ref39}, 3D MV Reg \cite{ref57} and BYOC \cite{ref13}, we also compare our method to the state-of-the-art point cloud registration method REGTR \cite{ref38}. 
 % REGTR is an end-to-end method trained on 3DMatch. 
 We use its officially provided weights, which are obtained by training on 3DMatch, for inference on ScanNet to compare the generalization. 
 The second category is only based on RGB image. 
 It includes many classic baselines such as SIFT \cite{ref5}, SuperPoint \cite{ref56}, and the recently proposed UR\&R \cite{ref12}. 
 To further verify the effectiveness of our method, we also incorporate a supervised version of UR\&R as our competitor. 
 The last category is based on RGB-D images. 
 We compare our method to the state-of-the-art method LLT \cite{ref14} and an RGB-D version of UR\&R, which treats depth information as an additional input channel.

 \subsection{Evaluation on ScanNet}\label{sec52}
 To fully evaluate the proposed method, we train our PointMBF on ScanNet \cite{ref21} and 3DMatch \cite{ref22}, respectively, and test them on ScanNet. 
 % The former experiment closely resembles the cases of processing unannotated datasets, while the latter evaluates the generalization of PointMBF. 
 The former experiment closely resembles the cases of processing unannotated datasets, while the latter evaluates the generalization.

 \noindent
 \textbf{Trained on ScanNet.} 
 % The results of models trained on ScanNet \cite{ref21} are shown in Table \ref{table1}. 
 % The performance of the proposed method and its competitors on the test set of ScanNet \cite{ref21} are shown in Table \ref{table1}, where all learning-based methods are also trained on ScanNet. 
 As shown in Table \ref{table1}, when the training set and test set come from the same domain, our proposed method achieves new state-of-the-art performances on almost all metrics, especially in terms of accuracy under small thresholds. 
 Compared to previous state-of-the-art method LLT \cite{ref14}, our method gains large improvement in translation, which is the bottleneck of the registration on ScanNet. 
 % Besides, our proposed method achieves lower chamfer errors, indicating our advantage on processing ambiguous scenes. 
 Moreover, by comparing our method with the RGB-D version of UR\&R and LLT, we find that the fusion strategy plays an important role in RGB-D point cloud registration. 
 % Treating deep information as an additional channel is an intuitive method, but has limited improvement. 
 Unidirectional fusion in LLT leverages the complementary information of RGB-D data, but still does not fully exploit them. 
 Our multi-scale bidirectional fusion is a better choice for RGB-D fusion, which can achieve better performance even without sophisticated branches as other methods \cite{ref14}. 
 This will be further demonstrated in our ablation studies. 
%  \vspace{-0.1em}

 \noindent
 \textbf{Trained on 3DMatch.} 
 % The results of the models trained on 3DMatch \cite{ref22} are also shown in Table \ref{table1}. 
 The generalization results of learning-based methods are also shown in Table \ref{table1}, where the models are trained on 3DMatch and tested on ScanNet. 
 It can be observed that our method not only achieves the state-of-the-art performance on almost all metrics but also outperforms several recent supervised methods such as REGTR \cite{ref38} and the supervised UR\&R by a large margin. 
 Overall, our method shows competitive generalization. 
 What is noticeable is that when the proposed method is trained on 3DMatch and tested on unseen ScanNet, its performance is comparable or even superior to LLT trained directly on ScanNet on some metrics.
 % It achieves comparable performance on rotation with LLT trained on ScanNet and even outperforms LLT trained on ScanNet when low errors are required. 
 % Similar to the above results on models trained on ScanNet, our multi-scale bidirectional fusion leverages complementary information more effectively than treating deep information as an additional channel or unidirectional fusion. 
 % Our following ablation studies will further verify the effectiveness of our fusion strategy. 
%  \vspace{-0.3em}

 \subsection{Ablation Studies}\label{sec53}
 To further verify the effectiveness of our multi-scale bidirectional fusion, we conduct comprehensive ablation studies. 
 All the models in our ablation studies are trained on 3DMatch \cite{ref22} and tested on ScanNet \cite{ref21}.

 \noindent
 \textbf{Comparison with other fusion strategies.} 
 As discussed in the related work section, there exist many other fusion strategies including undirected and unidirectional fusion. 
 To fully show the effectiveness of our fusion strategy, we compare our multi-scale bidirectional fusion with many other fusion strategies. 
 
 The results are shown in Table \ref{table2}. 
 It can be seen that our fusion strategy outperforms other strategies including undirected fusion (direct concatenation, DenseFusion \cite{ref37}) and unidirectional fusion (transformer-based fusion like DeepFusion \cite{ref23}), indicating the effectiveness of multi-scale bidirectional fusion. 
%  Besides, according to Table \ref{table1} and Table \ref{table2}, we find our visual branch performs worse than the visual branch in LLT \cite{ref14} and our geometric branch has poor performance, but our PointMBF still outperforms LLT. 
 Besides, Table \ref{table2} also shows the performance of the two single branches of PointMBF and the visual branch of LLT, and according to Table \ref{table1} and Table \ref{table2}, we find our visual branch performs worse than the visual branch of LLT \cite{ref14} in rotation and our geometric branch has poor performance, but our PointMBF still outperforms LLT. 
 These results strongly suggest that our multi-scale fusion can exploit the complementary information between different modalities more effectively. 
 We also find that the residue design in our fusion module plays an essential role. 
 This is because we use fusion in all stages, which tends to cause redundancy in the network. 
 Moreover, most fusion strategies successfully boost the performances, but fusion by treating depth information as an additional channel input causes performance degradation. 
 We speculate that the shared network can not deal with the big domain gap between RGB and depth. 
 It is more appropriate to use two different networks to handle different modalities. 
 This also reveals that the design of fusion strategy plays a vital role in RGB-D point cloud registration.

 \noindent
 \textbf{Effect of multi-scale fusion.} 
 In this work, we fuse information in all stages rather than in the last layers as LLT \cite{ref14}. 
 We believe that fusion in all stages can promote the exchange of complementary information in multiple scales, making features more distinctive. 
 % We believe that fusion in all stages can capture the semantics and geometric structures in multiply scales, making features more distinctive. 
 To verify this, we conduct an ablation on fusion stages. 
%  To avoid introducing too many extra hyperparameters, we coarsely divided fusion into three stages, encoding/decoding/concatenation, instead of considering whether to use fusion at each layer.
% \vspace{-0.1em} 

 The results are shown in Table \ref{table3}. 
 We find fusion in each stage all contributes to the feature extraction. 
 By gradually stacking fusion at different stages, our method finally achieves the best performance. 
 It also can be seen that our bidirectional fusion is powerful as only bidirectional fusion in the encoding or decoding stage shows competitive performance.

 \noindent
 \textbf{Effect of bidirectional fusion.} 
There are three types of information fusion in our proposed framework, namely the multi-scale visual-to-geometric (V2G) fusion, multi-scale geometric-to-visual fusion (G2V) fusion, and the fusion using direct concatenation (CAT) at the end of the two branches. 
To further confirm the effectiveness of each fusion, we conduct another ablation on fusion directions. 
Specifically, we reserve one or two of the three fusion types and compare their performance to our whole model. 
The results are shown in Table \ref{table4}. 
When only using one type of fusion, the performance of CAT is similar to that of G2V and they are superior to V2G. 
On base of CAT, adding either of V2G or G2V can help improve the performance and the highest performance is achieved by adding bidirectional fusion to CAT, as shown in the last row of Table \ref{table4}. 	

%  \vspace{-0.8em}
%  We have shown the superiority of our bidirectional fusion in above ablations. 
%  To further confirm its effectiveness, we conduct an ablation on fusion direction. 
%  Specifically, we only reserve one fusion direction in our network and test their performance. 
 
%  The results are shown in Table \ref{table4}. 
%  % We can find that both fusion from the geometric branch to the visual branch (G2V) and from the visual branch to the geometric branch (V2G) boost performance. 
%  We can find that both geometric-to-visual fusion (G2V) and visual-to-geometric fusion (V2G) boost performance. 
%  The network can achieve optimal performance only when fusing in both directions, which illustrates the effectiveness of the bidirectional fusion design. 
%  We also find that the latter direction i.e V2G improves the performance more. 
%  This is because $K_{v2g} > K_{g2v}$ makes more information embedded from RGB images to point clouds. 

 \section{Conclusion}
%  \vspace{-0.5em}
 In this work, we propose a multi-scale bidirectional fusion network for unsupervised RGB-D point cloud registration. 
 Different from other networks for RGB-D point cloud registration, our method implements bidirectional fusion in all stages rather than unidirectional fusion only at some stages, which can leverage the complementary information in RGB-D data more effectively. 
 The extensive experiments also show that our multi-scale bidirectional fusion not only helps network achieve new state-of-the-art performance but also outperforms a series of fusion strategies using the same network branches for feature extraction. 
 Furthermore, we believe our multi-scale bidirectional network is a general framework, which can be transferred to more applications such as reconstruction, tracking, etc in the future. 
 
 \section*{Acknowledgements} 
 This work was supported by the National Natural Science Foundation of China [grant number 62076070], and the Science and Technology Innovation Plan of Shanghai Science and Technology Commission [grant number 23S41900400].

{\small
\bibliographystyle{ieee_fullname}
\bibliography{egbib}
}

\clearpage

\appendix
In this supplementary material, we first list the details of our implementation in section \ref{s1}. 
 Second we conduct additional experiments including an analysis on our hyper-parameter, runtime analysis, and an evaluation on a more challenging dataset, in section \ref{s3}. 
Third, we provide the details of the network architecture in section \ref{s2}. 
Finally, we provide qualitative visualization of our PointMBF in section \ref{s4}.

\section{Implementation Details of PointMBF}\label{s1}
Table \ref{table5} shows the implementation details of our PointMBF. 
The first nine lines are the same as those in LLT \cite{ref14} and UR\&R \cite{ref12}, while the last three lines colored in grey are our own settings.

\begin{table}[hb]
  \centering
  \caption{ Implementation details of our PointMBF.}
  \scalebox{0.9}{
  \begin{tabular}{l|l}
  \hline
  Batch size                          & 8                            \\
  Image size                          & 128*128                      \\
  Feature dimension                   & 32                           \\
  Number of correspondence $k$          & 200                          \\
  Training epochs                      & 12                         \\
  Optimizer                           & Adam                         \\
  Learning rate                       & 1e-4                         \\
  Momentum                            & 0.9                          \\
  Weight decay                        & 1e-6                         \\
  \rowcolor[HTML]{EFEFEF} 
  $K_{v2g},K_{g2v}$ for training  & $K_{v2g}=16,K_{g2v}=1$ \\
  \rowcolor[HTML]{EFEFEF} 
  $K_{v2g},K_{g2v}$ for test  & $K_{v2g}=32,K_{g2v}=1$ \\
  \rowcolor[HTML]{EFEFEF} 
  Use pre-trained weight for ResNet18 & False                        \\ \hline
  \end{tabular}
  }
  \label{table5}
  \end{table}

\section{Additional Experiments}\label{s3}
In this section, we conduct two experiments including analysis on hyperparameter $K_{v2g}$ and runtime analysis. 
All the models in the experiments is trained on 3DMatch \cite{ref22} and tested on ScanNet \cite{ref21}.

\subsection{Effect of hyperparameter $K_{v2g}$}
Hyperparameter $K_{v2g}$ denotes the number of visual features embedded into each geometric feature. 
Here, we test its influence on registration. Limited by memory, we set it from 1 to 32. 

The result is shown in Table \ref{table6}. 
It can be seen that the performance of our method improves as $K_{v2g}$ increases, but the trend of improvement gradually slows down. 
This is because that, as $K_{v2g}$ increases, $K_{v2g}$ gradually reaches the number of points or pixels in a corresponding region. 
We also find that even if we set $K_{v2g}$ to 1, our method still outperforms the state-of-the-art method, LLT \cite{ref14} in almost all metrics, which illustrates the effectiveness of our method.

\subsection{Runtime Analysis}
We conduct runtime analysis by comparing time overhead on each step of unsupervised RGB-D registration. 
Both our PointMBF and the competitor, UR\&R are tested on an A40 graphic card with an Intel Xeon Platinum 8358P CPU. 
We report the mean and standard deviation of running time of each step. 

The result is shown in Table \ref{table7}. 
It can be seen that our multi-scale bidirectional fusion greatly improves performance by a large margin without adding much time overhead ($<10ms$). 
Furthermore, the extra overhead on feature extraction is negligible compared to the overhead of correspondence estimation (main overhead).

\subsection{Evaluation on ScanNet-SuperGlue}
 Our experiments in the main paper follow the settings of UR\&R \cite{ref12},  in which view pairs are 20 frames apart. 
 We find 20 frames apart is less challenging, making the effectiveness of our method less obvious. 
 Specifically, over 99.8\% of the ground truth rotation is under 45°, which makes ICP performs best under 45° threshold in Table \ref{table1}. 
 Moreover, too many easy cases also cause similar medians in Table \ref{table1}, especially for chamfer errors. 
 Therefore, we conduct a more challenging evaluation on ScanNet-SuperGlue.

 ScanNet-SuperGlue is a dataset based on ScanNet \cite{ref21}, which is provided by SuperGlue \cite{superglue}. 
 It includes 1500 view pairs with average 480.8 frames apart. Our competitors include UR\&R and a fusion-based method named CAT. 
 CAT utilizes the same visual and geometric branches as our method and fuses visual features and geometric features using a concatenation operator at the final stage. 
 All the above methods including our PointMBF are trained on 3DMatch \cite{ref22} and tested on ScanNet-SuperGlue.

 As shown in Table \ref{table8}, our method outperforms the others by a large margin. 
 Moreover, the median chamfer errors vary considerably because this experiment includes more hard cases.

% \usepackage{multirow}
\begin{table*}[ht]
 \centering
 \caption{Registration results under different hyperparameter $K_{v2g}$.}
 \scalebox{0.9}{
 \begin{tabular}{lccccccccccccccc}
 \hline
 \multirow{3}{*}{} & \multicolumn{5}{c}{Rotation (deg)}                                          & \multicolumn{5}{c}{Translation (cm)}                                        & \multicolumn{5}{c}{Chamfer (mm)}                                            \\
                   & \multicolumn{3}{c}{Accuracy↑}                 & \multicolumn{2}{c}{Error↓}  & \multicolumn{3}{c}{Accuracy↑}                 & \multicolumn{2}{c}{Error↓}  & \multicolumn{3}{c}{Accuracy↑}                 & \multicolumn{2}{c}{Error↓}  \\ \cline{2-16} 
                   & 5            & 10           & 45           & Mean         & Med.         & 5             & 10            & 25            & Mean         & Med.         & 1             & 5             & 10            & Mean         & Med.         \\ \hline
 LLT \cite{ref14}      & 93.4          & 96.5          & \textbf{98.8} & \textbf{2.5} & \textbf{0.8} & 76.9          & 90.2          & 96.7          & \textbf{5.5} & 2.2          & 86.4          & 95.1          & 96.8          & \textbf{4.6} & \textbf{0.1} \\ \hline
 $K_{v2g}$=1              & 93.7           & 96.6           & 98.6           & 3.2       & \textbf{0.8} & 79.5             & 91.1        & 96.7             & 6.6          & 2.2          & 90.2           & 96.1           & 97.2             & 5.4            & \textbf{0.1} \\
 $K_{v2g}$=2              & 93.7          & 96.4          & 98.7          & 3.2          & \textbf{0.8} & 79.7          & 91.2          & 96.6          & 6.5          & 2.2          & 90.3          & 96.1          & 97.1          & 5.3          & \textbf{0.1} \\
 $K_{v2g}$=4              & 94.0          & 96.6          & 98.6          & 3.2          & \textbf{0.8} & 80.1          & 91.6          & 96.8          & 6.6          & \textbf{2.1} & 90.6          & 96.3          & 97.2          & 5.4          & \textbf{0.1} \\
 $K_{v2g}$=8              & 94.3          & 96.7          & 98.7          & 3.1          & \textbf{0.8} & 80.4          & 91.6          & 96.9          & 6.5          & \textbf{2.1} & 90.9          & 96.4          & 97.3          & 5.2          & \textbf{0.1} \\
 $K_{v2g}$=16             & 94.5          & 96.9          & 98.7          & 3.1          & \textbf{0.8} & 80.8          & 91.7          & 97.0          & 6.3          & \textbf{2.1} & 91.1          & 96.4          & 97.3          & 5.1          & \textbf{0.1} \\
 $K_{v2g}$=32             & \textbf{94.6} & \textbf{97.0} & 98.7          & 3.0          & \textbf{0.8} & \textbf{81.0} & \textbf{92.0} & \textbf{97.1} & 6.2          & \textbf{2.1} & \textbf{91.3} & \textbf{96.6} & \textbf{97.4} & 4.9          & \textbf{0.1} \\ \hline
 \end{tabular}
 }
 \label{table6}
\end{table*}

\begin{table*}[ht]
 \centering
 \caption{
     Performance on ScanNet (Splitted by SuperGlue). CAT denotes fusion using direct concatenation.
 }
 \scalebox{0.9}{
 \setlength{\tabcolsep}{3mm}{  
    \begin{tabular}{lcccccccccccc}
       \hline
       \multirow{3}{*}{} & \multicolumn{4}{c}{Rotation (deg)}                           & \multicolumn{4}{c}{Translation (cm)}                          & \multicolumn{4}{c}{Chamfer (mm)}                             \\
                         & \multicolumn{3}{c}{Accuracy↑}                 & Error↓       & \multicolumn{3}{c}{Accuracy↑}                 & Error↓        & \multicolumn{3}{c}{Accuracy↑}                 & Error↓       \\ \cline{2-13} 
                                           & 5             & 10            & 45            & Med.         & 5             & 10            & 25            & Med.          & 1             & 5             & 10            & Med.         \\ \hline
       UR\&R  \cite{ref12}           & 36.0          & 49.0          & 82.3          & 10.5         & 18.5          & 29.7         & 48.3         & 26.9          & 24.6         & 41.3          & 48.8         & 11.1        \\
       CAT               & 47.1          & 56.8          & 81.1          & 6.2          & 27.7          & 41.2          & 55.5          & 17.7          & 40.5          & 54.1          & 59.7          & 3.2          \\
       Ours              & \textbf{51.1} & \textbf{60.8} & \textbf{82.9} & \textbf{4.7} & \textbf{31.5} & \textbf{44.2} & \textbf{59.3} & \textbf{13.8} & \textbf{44.8} & \textbf{57.5} & \textbf{63.3} & \textbf{1.7} \\ \hline
       \end{tabular}
 }
 }
 \label{table8}
\end{table*}

\begin{table}[]
 \centering
 \caption{Runtime analysis.}
 \scalebox{0.86}{
 \begin{tabular}{l|cc}
 \hline
 \multirow{2}{*}{}             & \multicolumn{2}{c}{Time (ms)}                      \\ \cline{2-3} 
                               & \multicolumn{1}{c|}{UR\&R \cite{ref12}} & Ours         \\ \hline
 Feature Extraction            & \multicolumn{1}{c|}{21.94±20.92}     &  31.92±8.97 \\
 Correspondence Estimation     & \multicolumn{1}{c|}{247.88±23.22}   & 255.94±22.95 \\
 Geometric Fitting             & \multicolumn{1}{c|}{9.11±5.25}      & 8.91±5.44    \\
 Rendering (Just for training) & \multicolumn{1}{c|}{8.32±6.73}      & 8.10±6.28    \\ \hline
 \end{tabular}
 }
 \label{table7}
\end{table}

\section{Details of the Network Architecture}\label{s2}
In this section, we provide details of the network architecture including feature extractor, point/pixel gathering for fusion, geometric fitting, keypoint, and differentiable rendering.

\subsection{Feature extractor}
Figure \ref{fig4} shows the detailed architecture of the feature extractor in our PointMBF. 
As mentioned in section \ref{sec31}, we modify a ResNet18 \cite{ref19} into our visual branch. 
The encoder consists of conv2\_x, conv3\_x, and conv4\_x in ResNet18, and we remove the max pooling in original conv2\_x. 
The decoder mainly consists of upsampling module, concatenation operator, and convblock i.e. shallow perception module.  
Our geometric branch has a symmetric architecture to the visual branch.

\subsection{Point/pixel gathering for fusion}
\begin{figure*}[ht]
   \begin{center}
   \centerline{\includegraphics[width=2.0\columnwidth]{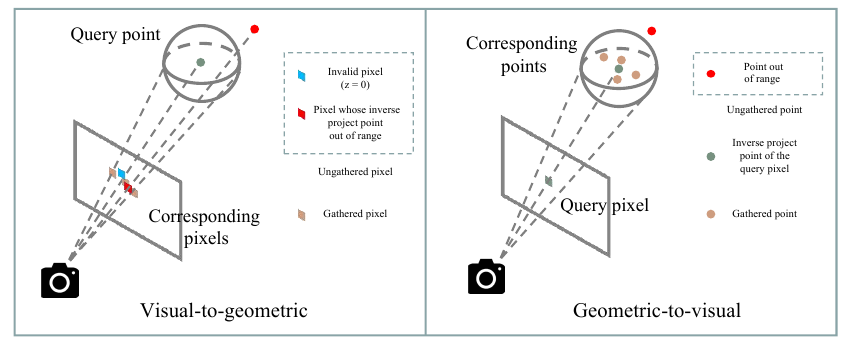}}
   \caption{
     Point/pixel gathering for fusion.
     }
   \label{fig9}
   \end{center}
 \end{figure*}

   During our feature extraction, we embed regional visual features into geometric features and regional geometric features into visual features. 
   In this procedure, it’s important to gather corresponding points/pixels for fusion. 
   Here, we provide the details of the point/pixel gathering process, which is shown in Figure \ref{fig9}.

   For visual-to-geometric fusion in the $l$-th layer, given a query point, we first determine its neighbor ball with radius $R^l_{v2g}$. 
   Then we project this neighbor ball to the camera and the pixels falling in the projection region are selected as candidate pixels for fusion. 
   After that, we filter the improper pixels in candidate pixels.
   There are two categories of inproper pixels. 
   The first category is the invalid pixel, whose corresponding depth z is zero. 
   These pixels may represent noise or holes in depth images. 
   Embedding features from invalid pixels may deliver improper information. 
   The second category is the pixel, whose inverse project point is out of the neighbor ball of the query point. 
   Points outside of the query point’s neighbor ball may also be projected to pixels that are close to the projection of the query point, but these points are less related to the query point in 3D semantics, so they are also filter out. 
   Finally, we gather the pixels within remaining pixels for fusion, whose inverse project points are in the K nearest neighbors of the query point.

   For geometric-to-visual fusion in the $l$-th layer, given a query pixel, we first determine its inverse project point. 
   Then the points which fall in the neighbor of the inverse project point with radius $R^l_{g2v}$ are selected as candidate points. 
   Finally, we gather the KNN points of the inverse project point within the candidate points.

\subsection{Keypoint}
 In this work, we utilize dense descriptions for correspondence estimation. 
 In other words, all the points in the point clouds are considered as keypoints except invalid points with depth z=0.

\subsection{Geometric fitting}
   Our geometric fitting is the same as that in UR\&R \cite{ref12}, which is a modified RANSAC \cite{ref6}. 
   We provide its detail for convenience in this section.

 Given 400 input correspondences $C=\left\{\left(p^\mathcal{S}, p^\mathcal{T}, w\right)_i: 0 \leq i<2 k\right\}$ with their corresponding weights, we randomly sample $t$ subsets of $C$ and estimate $t$ candidate transformations. 
 Each subset contains $l$ randomly sampled correspondences, and each candidate transformation is estimated by solving a weighted Procrustes problem \cite{ref7} on a subset. 
 Then we choose the candidate transformation $T^*$ with minimal error $E(C,T^*)$ in equation \ref{eq7} as our final estimation. 
 During training, we set $t$ to 10 and $l$ to 80. At test time, we set $t$ to 100 and $l$ to 20.

\subsection{Differentiable rendering}
\begin{figure}[ht]
 \begin{center}
 \centerline{\includegraphics[width=1.0\columnwidth]{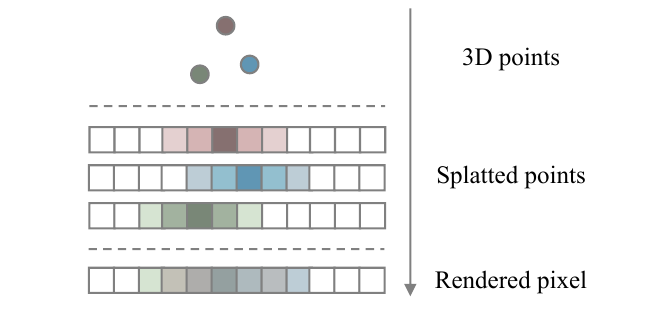}}
 \caption{
   The basic principle of differentiable rendering. 
   Differentiable rendering softens the process of projection. 
   Each 3D point affects a certain region of pixels by splatting itself to a region, and the rendered pixels are the accumulation of all the splatted points.}
 \label{fig8}
 \end{center}
\end{figure}

\begin{figure*}[ht]
  \begin{center}
  \centerline{\includegraphics[width=2.0\columnwidth]{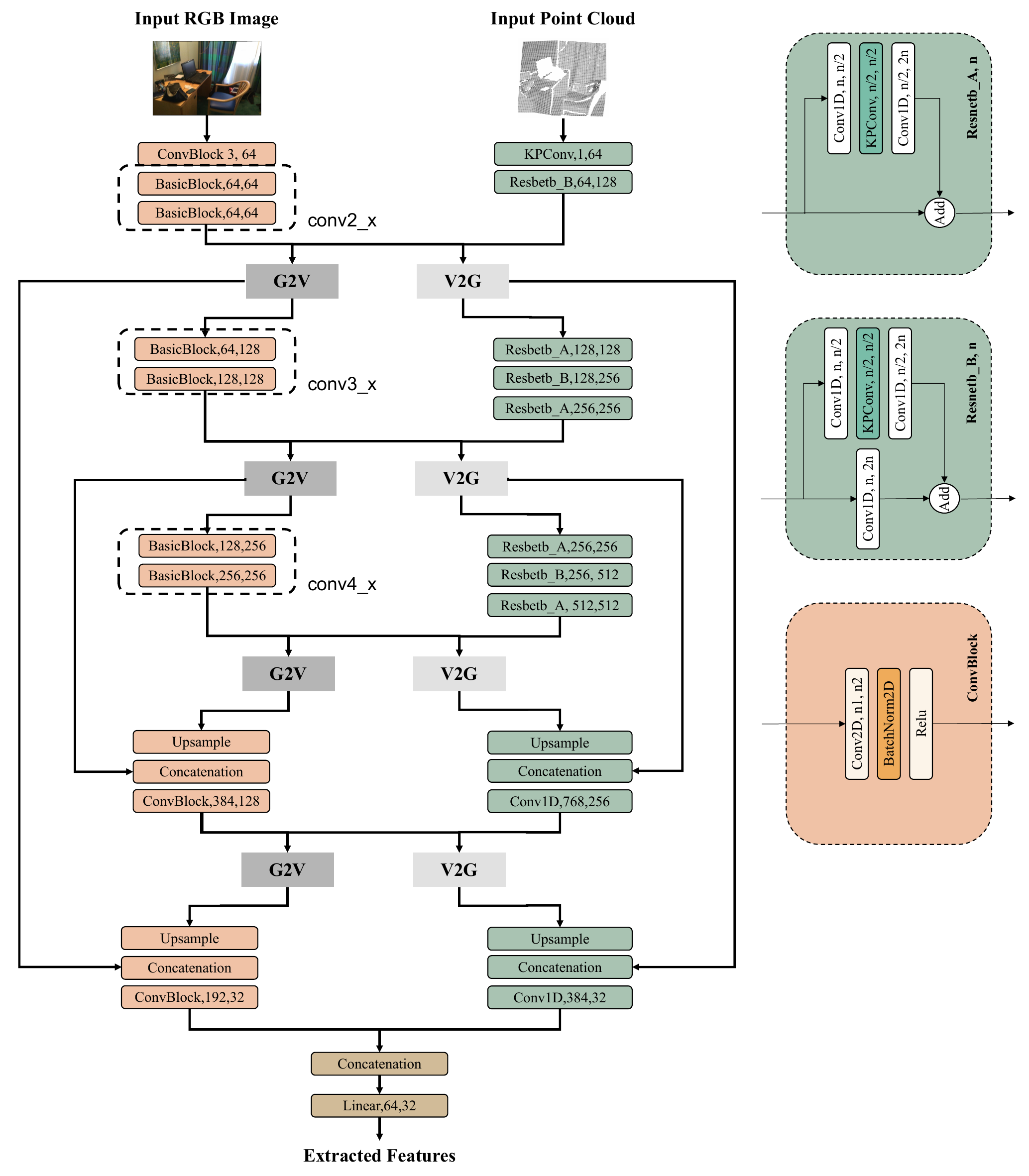}}
  \caption{
    The detailed architecture of our feature extractor.
    }
  \label{fig4}
  \end{center}
\end{figure*}

 The differentiable renderer is a rendering technique that leverages differentiable programming to optimize and compute gradients of the rendering process. 
 Its basic principle is shown in Figure \ref{fig8}. 
 It softens the process of projection, whereby each pixel is the accumulation of multiple splatted points. 
 This allows each point to receive gradients from multiple pixels, avoiding the local gradient \cite{render} issues caused by hard rasterization. 
 Furthermore, the accumulation strategy employed in the soft projection approximates the occlusion observed in the natural world. 
 We implement our differentiable renderer using Pytorch3d \cite{ref53}. 
 It takes transformed point clouds as input and outputs rendered images for photometric loss calculation.

\section{Qualitative Visualization}\label{s4}
We provide detailed visualization in this section. 
We visualize the inputs, extracted features, generated correspondences, and final registration results in two challenging scenes, including cluttered and ambiguous scenes.

\begin{figure*}[ht]
  \begin{center}
  \centerline{\includegraphics[width=2.0\columnwidth]{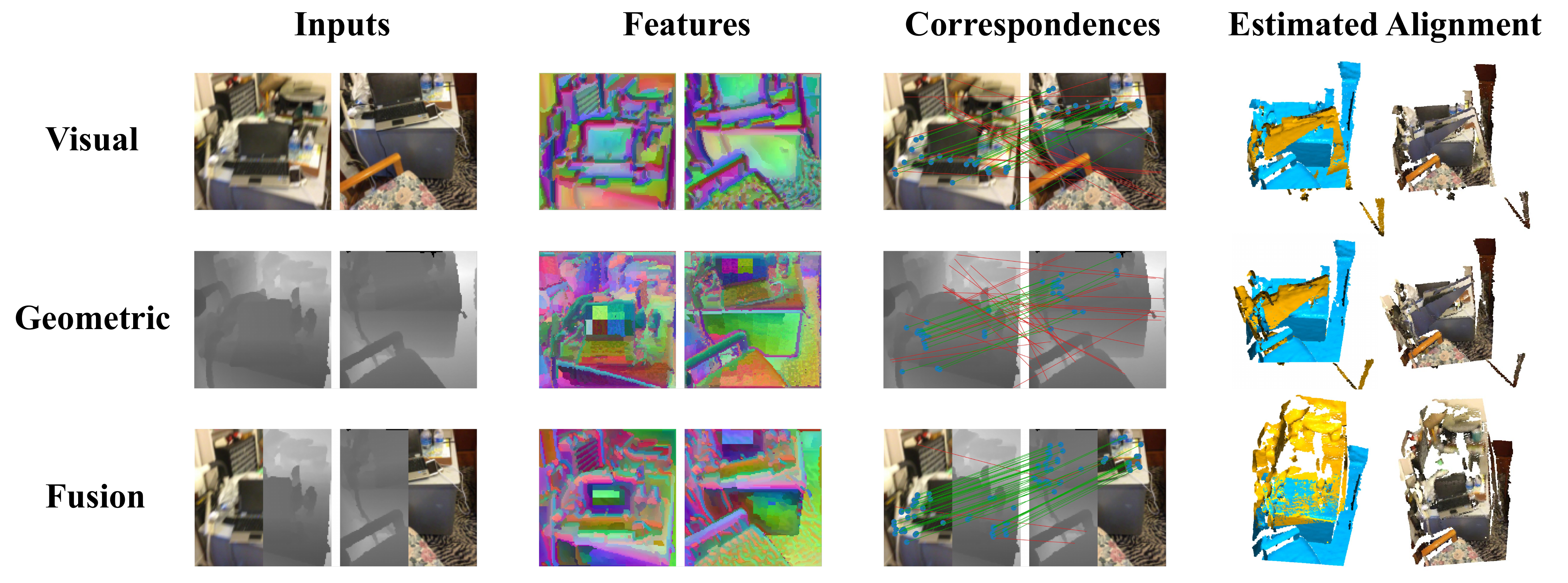}}
  \caption{
    \textbf{Visualization on RGB-D registration in cluttered scene. }
    The features are visualized by mapping them to colors by t-SNE \cite{ref58}. 
    The red lines denote the outlier correspondences, while the green lines denote the inlier correspondences.
    }
  \label{fig5}
  \end{center}
\end{figure*}

The results of the cluttered scene are shown in Figure \ref{fig5}. 
The first two rows show the registration of two single branches, and the last row shows ours. 
It can be seen that in a cluttered scene, there exist complex semantics, partial overlap, and blur caused by camera jitter, which make registration challenging. 
Visual and geometric branches can not deal with it perfectly and tend to generate more outlier correspondences, leading to registration failure. 
But our method considers both semantics and local geometry, tends to avoid wrong correspondences.

\begin{figure*}[ht]
  \begin{center}
  \centerline{\includegraphics[width=2.0\columnwidth]{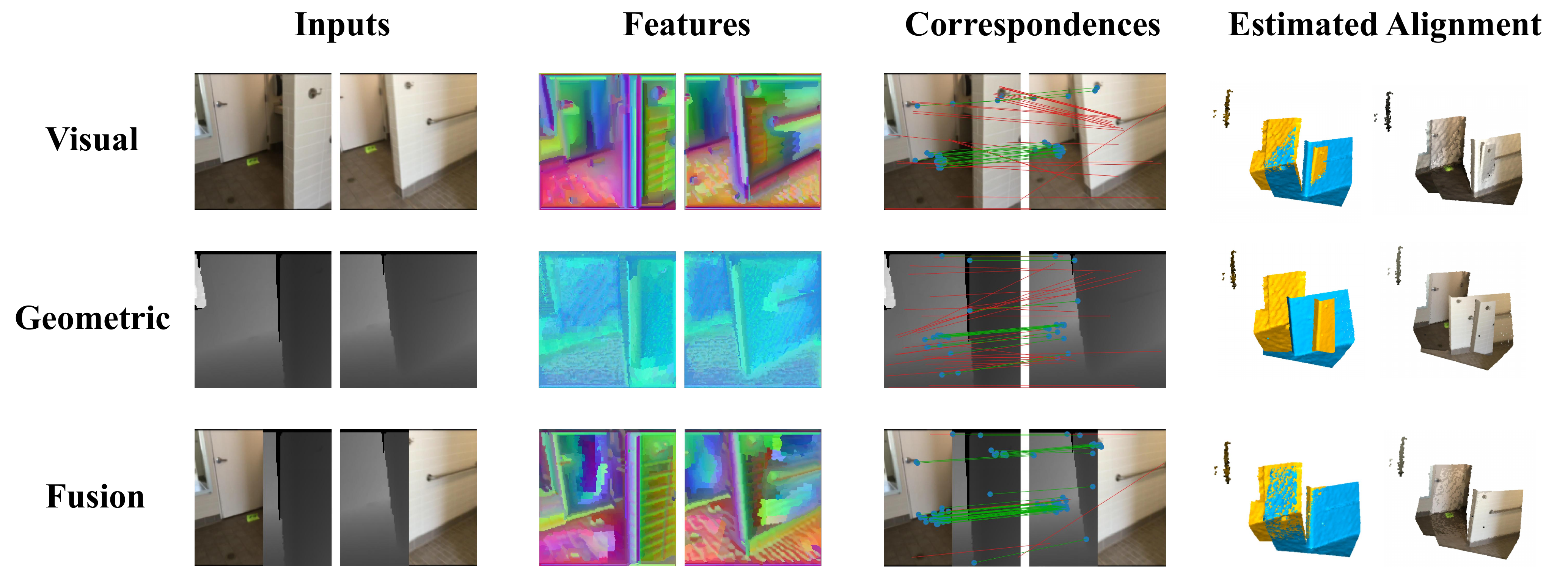}}
  \caption{
    \textbf{Visualization on RGB-D registration in ambiguous scene. }
    The features are visualized by mapping them to colors by t-SNE \cite{ref58}. 
    The red lines denote the outlier correspondences, while the green lines denote the inlier correspondences.
    }
  \label{fig6}
  \end{center}
\end{figure*}

\begin{figure*}[ht]
  \begin{center}
  \centerline{\includegraphics[width=2.0\columnwidth]{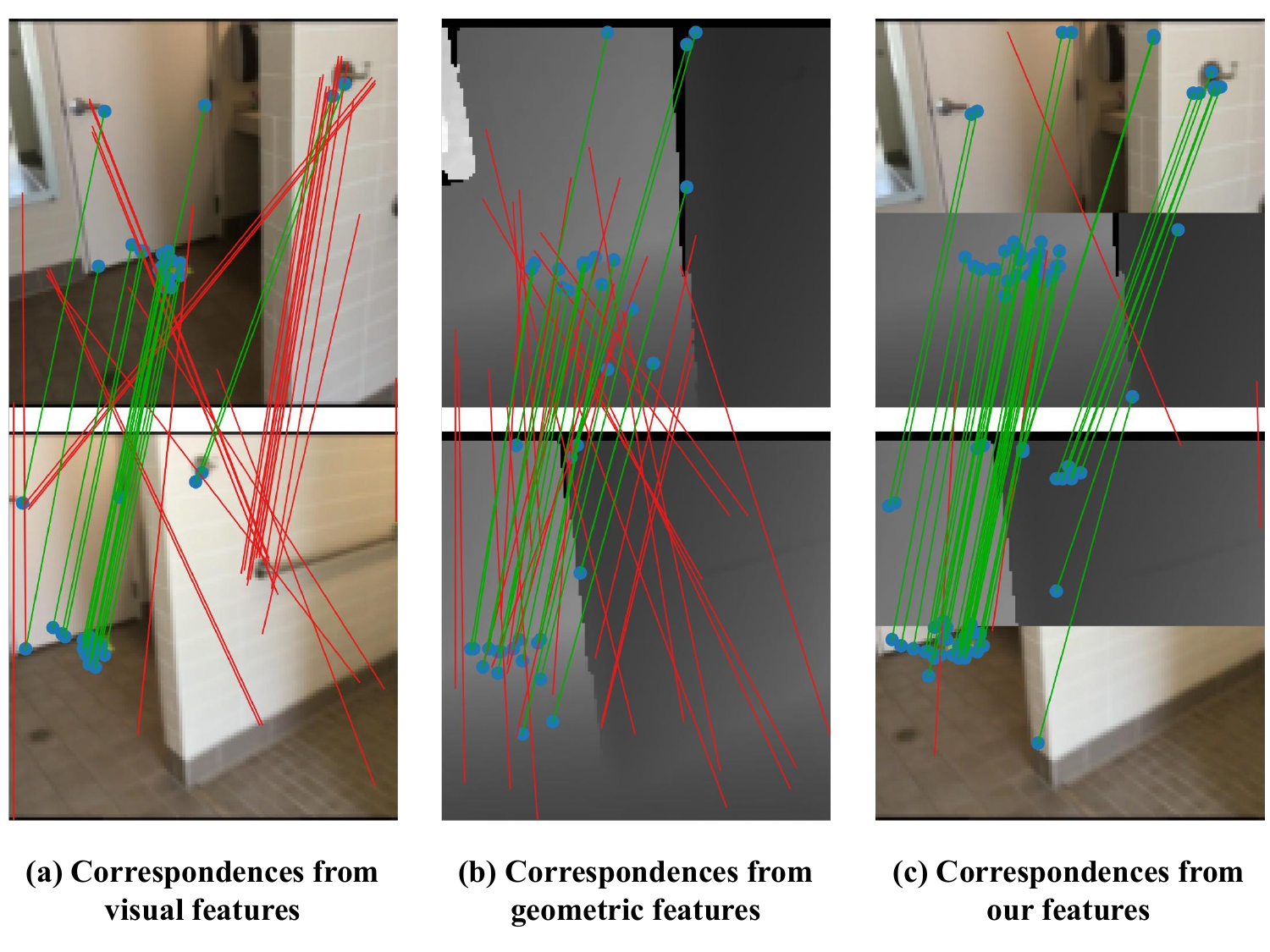}}
  \caption{
    \textbf{Zoom-in visualization for correspondences in Figure \ref{fig6}. }
    The red lines denote the outlier correspondences, while the green lines denote the inlier correspondences. 
    The visual features can not distinguish the hook from the armrest and the geometric features confuse the floor with the wall. 
    But our fused features can generate reliable correspondences for registration as they consider the complementary information from RGB-D data.
    }
  \label{fig7}
  \end{center}
\end{figure*}

The results of the ambiguous scene are shown in Figure \ref{fig6}. 
The first two rows show the registration of two single branches, and the last row shows ours. 
It can be seen that in an ambiguous scene, there exist many ambiguous and repetitive structures such as floors, walls, and symmetrical objects without textures, making correspondences based on a single modality contain a large proportion of outliers. 
Our fusion considers both semantics from RGB information and local geometric distributions from point clouds, which greatly improves the performance. 
For example, as shown in Figure \ref{fig7}, the visual features can not distinguish the hook from the armrest due to similar local texture and the geometric features produce more wrong correspondences because of too many repetitive surfaces in this scene. 
However, our fused features successfully distinguish them and produce correspondences from a more reliable area. 
\vspace{-50pt}

\end{document}